%% file: Main.tex
\documentclass{article} 
\usepackage{iclr2026_conference,times}
\iclrfinalcopy

\usepackage{authblk} 
\usepackage{hyperref}
\usepackage{url}
\usepackage{graphicx}
\usepackage{booktabs} 
\usepackage{tabularx}
\usepackage{rotating}
\usepackage{longtable}
\usepackage{subcaption}

\title{Rare Event Early Detection: A Dataset of Sepsis Onset for Critically Ill Trauma Patients}

\author{
\begin{tabular}{c}
\textbf{Yin Jin}$^{1}$ \quad
\textbf{Tucker R. Stewart}$^{1}$ \quad
\textbf{Deyi Zhou}$^{1}$ \quad
\textbf{Chhavi Gupta}$^{1}$ \quad
\textbf{Arjita Nema}$^{2}$ \quad
\textbf{Scott C. Brakenridge}$^{3}$ \quad
\textbf{Grant E. O'Keefe}$^{3}$ \quad
\textbf{Juhua Hu}$^{1}$ \quad
\\ 
$^{1}$School of Engineering and Technology, University of Washington, Tacoma, WA, USA \\
$^{2}$Department of Statistics, University of Washington, Seattle, WA, USA \\
$^{3}$Department of Surgery, University of Washington, Seattle, WA, USA
\\
\texttt{\{jinyin,trstew,oliviady,cgupta4,arjita28,sbrakenr,gokeefe,juhuah\}@uw.edu} 
\end{tabular}
}

\begin{document}

\maketitle

\begin{abstract}
Sepsis is a major public health concern due to its high morbidity, mortality, and cost. Its clinical outcome can be substantially improved through early detection and timely intervention. By leveraging publicly available datasets, machine learning (ML) has driven advances in both research and clinical practice. However, existing public datasets consider ICU patients (Intensive Care Unit) as a uniform group and neglect the potential challenges presented by critically ill trauma patients in whom injury-related inflammation and organ dysfunction can overlap with the clinical features of sepsis. We propose that a targeted identification of post-traumatic sepsis is necessary in order to develop methods for early detection. Therefore, we introduce a publicly available standardized post-trauma sepsis onset dataset extracted, relabeled using standardized post-trauma clinical facts, and validated from MIMIC-III. Furthermore, we frame early detection of post-trauma sepsis onset according to clinical workflow in ICUs in a daily basis resulting in a new rare event detection problem. We then establish a general benchmark through comprehensive experiments, which shows the necessity of further advancements using this new dataset. The data code is available at \href{https://github.com/ML4UWHealth/SepsisOnset_TraumaCohort.git}{https://github.com/ML4UWHealth/SepsisOnset\_TraumaCohort.git}.
\end{abstract}

\section{Introduction}
\input{S1_Intro}

\section{Related Work}
In this section, we briefly review related work from two perspectives, that is, sepsis-related datasets and ML techniques for early sepsis detection and rare event.
\input{S2_RW}

\section{Dataset Construction}
Using the MIMIC-III dataset \citep{johnson2016mimic}, an open-access anonymized database of 61,532 admissions from 2001 to 2012 across six ICUs, our work aims to construct a public dataset to facilitate early detection of post-traumatic sepsis. Our approach differs from previous work using this dataset by focusing specifically on post-traumatic sepsis that consists of three modules: 1) standardizing a trauma-focused cohort, 2) adopting carefully designed sepsis criteria targeting trauma patients based on the MIMIC-III dataset, and 3) introducing a clinically aligned detection framework. The selection of critical criteria will be defined, explained, and referenced in this work to ensure data quality.

\input{S3_Dataset_icml}

\section{Benchmark Task \& Proposed Method}
\input{S4_Method}

\section{Experiments}
\input{S5_EXP_icml}

\section{Conclusions}
Considering that no public data are available for post-trauma sepsis in ICUs that is different from other ICU sepsis and harder to detect but critical to morbidity and mortality after severe
traumatic injury, this work constructs a standardized trauma cohort with reliable and reproducible post-traumatic sepsis onset labels. Comprehensive experiments show promising advances using the new dataset in both post-traumatic sepsis and rare event early detection, which also shows the importance of reliable labels even on a relatively small dataset. In the future, the same pipeline can be extended to other databases (e.g., MIMIC-IV, eICU) to further improve the generalizability of our dataset.

\bibliography{uwSepsis}
\bibliographystyle{iclr2026_conference}

\clearpage
\appendix
\section*{Appendix}
\input{Appendix_ICML}

\end{document}

%% file: S1_Intro.tex
Sepsis is a life-threatening condition associated with an altered immune response to infection and remains a major public health concern with high morbidity, mortality, and economic burden \citep{singer2016third, fleischmann2016assessment, rudd2020global}. By leveraging electronic health records (EHRs) and advanced ML techniques for sepsis management, the AI Clinician~\citep{komorowski2018artificial} pioneered a ML-driven optimization for sepsis treatment by generating sepsis cohorts from public ICU datasets.  More recently, PhysioNet launched a challenge focusing on the early detection of sepsis, attracting wide participation from academia and industry~\citep{reyna2020early}. These efforts demonstrate how public datasets have not only facilitated ML research on sepsis and may have a potential impact on clinical care. For example, the Targeted Real-Time Early Warning System (TREWS) showed improved sepsis management with ML~\citep{adams2022prospective}.

However, almost all previous work ignored the special fact that sepsis remains a prevalent complication and a significant contributor to morbidity and mortality after severe traumatic injury \citep{guirgis:2016-long, stern2023defining}. This could be caused by a delay in the detection of post-traumatic sepsis. This delay is, in part, because critically ill trauma patients often exhibit physiological responses and alterations in organ function (e.g., acute lung injury and acute kidney injury) that are due to the initial trauma and exist before the development of infection~\citep{eriksson:2019-comparison, eguia2020trends, stern2023defining}. This overlap makes the trauma cohort different from other critically ill patients and may hinder the ability to detect sepsis early. For example, AI Clinician~\citep{komorowski2018artificial} and  PhysioNet Challenge~\citep{reyna2020early} both focus on the general ICU population, and provide sepsis labels based on the standard Sepsis-3 definition~\citep{singer2016third}. For convenience, research groups studying specific trauma cohorts~\citep{fu2019early, guo2024developing} directly use readily available sepsis-3 labels, without incorporating trauma-specific clinical guidance. As a result, current ML researchers often rely on datasets that do not accurately reflect the realities of post-trauma physiology and clinical care.

To address this gap, in this work, we introduce a publicly available standardized post-trauma sepsis onset dataset that is extracted, relabeled, and validated from MIMIC-III, a large database with more than 40,000 patients who stayed in critical care units. Specifically, our dataset is the first to provide a well-defined standardized trauma cohort with reliable post-traumatic sepsis onset labels. The dataset is supported by a well-documented reproducible code base for future research and clinical applications.
In addition, to align with clinical practice in the ICU, we frame the early detection of sepsis on a daily basis for each patient, for whom evaluation and treatment planning occurs primarily during morning rounds~\citep{stewart2023nprl}. Specifically, instead of labeling each patient's entire visit to have sepsis (i.e., between 13.6\% and 39.3\% in the ICU) or not, each day within the visit is examined for the first possible occurrence of sepsis onset~\citep{stewart2023nprl}. Thereafter, sepsis examples become rare compared to negative examples, resulting in a severe class imbalance problem.  
It should be noted that early detection of rare events is a fundamental challenge in ML \cite{shyalika2024comprehensive}, not limited to sepsis, making our dataset a valuable benchmark for developing and evaluating such methods. To support this research, we conducted a comprehensive empirical study by extending existing ML techniques and established a general benchmark on this new rare event early detection problem, highlighting future opportunities for improvement. Our contributions can be summarized as follows:

\begin{itemize}
    \item We are the first to emphasize the difference between general ICU sepsis and post-traumatic sepsis, and to build a standardized trauma cohort with targeted post-traumatic sepsis onset labels to facilitate its early detection.
    \item We frame early detection of post-traumatic sepsis onset according to real clinical practice in the ICU, which has been ignored before, although resulting in a challenging rare event problem.
    \item We conducted comprehensive experiments by extending existing ML techniques and established a general benchmark for early detection of post-traumatic sepsis onset, showing future opportunities using this new dataset.
\end{itemize}

%% file: S2_RW.tex
\subsection{Publicly Available Sepsis Definition}
For a sepsis dataset, it is critical to identify sepsis patients and the timestamp of sepsis onset. The Third International Consensus Definitions for Sepsis and Septic Shock (Sepsis-3)~\citep{singer2016third} is widely used and is the current consensus definition of sepsis in the ML community. AI Clinician \citep{komorowski2018artificial} pioneered the use of machine learning for sepsis-relevant tasks. By leveraging public datasets such as MIMIC-III \citep{johnson2016mimic} and eICU \citep{pollard2018eicu}, AI Clinician provides a structured framework for sepsis definitions based on the Sepsis-3 criteria. Publicly available datasets and their corresponding open-source code bases ensure the reproducibility of Sepsis-3 and facilitate future sepsis-related research. Later on, the PhysioNet Challenge 2019 \citep{reyna2020early} introduced a standardized benchmark for early sepsis detection. Although it provides sepsis labels for each patient along with documentation on how sepsis was defined, the reproducibility of sepsis labels for future use remains limited. In addition, the official MIMIC code repository has recently released SQL implementations of Sepsis-3~\citep{mimic_code}, further supporting reproducibility and consistency in sepsis-related studies.

Although all above public datasets follow the general Sepsis-3 guideline, describing sepsis as life-threatening organ dysfunction resulting from a dysregulated host response to infection, their detailed criteria are not identical. For example, AI Clinician identifies organ dysfunction when the SOFA score is greater than 2, while PhysioNet defines it based on a SOFA score increase of 2~\citep{johnson2018comparative}. Moreover, PhysioNet enforced stricter feature extraction by emphasizing blood cultures and intravenous antibiotic routes. As a result, AI Clinician tends to identify a larger, less critically ill cohort, resulting in higher false positive sepsis rates. While PhysioNet improves upon this, its reproducibility and clarity is limited without open source code support.

Beyond the variations in how Sepsis-3 criteria are operated across studies, another limitation is that the Sepsis-3 definition itself does not fully address the unique characteristics of trauma patients. 
\citep{stern2023defining} provides a framework to computationally define sepsis in trauma patient databases. However, no publicly released dataset has incorporated this more specific definition. Given the high morbidity and mortality of hospital-acquired sepsis among trauma patients, recent studies have begun exploring sepsis detection in this subgroup. However, they either rely on private datasets \citep{stewart2023nprl, 10487386, guo2024developing} or use public datasets with different cohort extraction criteria, and lack open source code support~\citep{fu2019early, li2023real}. This severely restricts the reproducibility of trauma-specific studies. In this work, we adopt a refined definition of trauma-specific sepsis on a public dataset with a reproducible code base, ensuring a standardized foundation for future research of post-traumatic sepsis. 

\subsection{Early Sepsis Detection}
Early sepsis detection involves two main challenges: deriving meaningful physiological representations from raw, noisy time-series data within an observation window and distinguishing early sign of sepsis cases from all cases based on the given features, which are experimented in this work. 

Feature engineering is a straightforward approach where handcrafted statistical descriptors such as mean, maximum, or minimum are computed over each vital sign within the observation window~\citep{9005773, 10487386}. Such features provide a coarse summary of the patient’s physiological state but fail to capture temporal dynamics. To address this limitation, \citep{9005805} introduces mathematical tools such as the signature transformation (also known as path signature), which converts multivariate time series into fixed-length feature vectors that preserve temporal information. More recently, deep learning models have been widely adopted to automatically extract hidden temporal features, with commonly used architectures including recurrent networks such as gated recurrent units(GRUs)~\citep{9005773, 9005856} and Long-Short-Term Memory(LSTMs)~\citep{9005879, 9629559, stewart2023nprl, 10783411}, as well as temporal convolutional neural networks (TCNNs)~\citep{lauritsen2020explainable, 10701306}. These deep learning models have greater potential to learn complex features but with a high demand of training data. 

Given the feature representations, the classification can be broadly grouped into two categories: tree-based models and fully connected layer (FCL) classification heads. Tree-based models, such as XGBoost and LightGBM, have been widely applied in early sepsis detection~\citep{9005805, 10487386}. They are friendly to small datasets and offer fast training, but their performance heavily depends on the quality of the input features, and they are limited in capturing complex, high-dimensional relationships. In contrast, FCL classification heads are typically used on top of deep representation models~\citep{9005565, stewart2023nprl}. They possess a stronger expressive power, allowing them to capture non-linear interactions.

\subsection{Imbalance and Rare Event Challenge}
Early sepsis detection faces an imbalance challenge or even a rare event challenge. Most existing work applied data resampling techniques, such as oversampling and undersampling \citep{lauritsen2020early, teredesai2022sub, ewig2023multi}. However, these methods often lead to overfitting, in the case of oversampling, or loss of critical information, in the case of undersampling. In contrast to this, reweighting-based strategies  alleviate the imbalance by assigning different loss weights to positive and negative samples \citep{10783411, 10731539}. However, reweighting struggles with highly imbalanced datasets as extreme weight differences can lead to unstable training and overfitting to the minority class. 

Therefore, conventional imbalance handling methods remain insufficient, especially in rare event detection tasks like ours, where the minority class not only constitutes a small fraction, but also has a limited number of examples. \citet{9629559} proposed to first apply a Variational autoencoder (VAE) to learn feature representations, and then use a clustering algorithm to classify into normal and anomalous groups. Although such methods can help identify rare events, they rely on the assumption that anomaly examples have a distinct distribution. In our case, trauma patients with sepsis often exhibit physiological patterns similar to general trauma patients, making simple anomaly detection less effective. Alternatively, \citep{stewart2023nprl} utilizes self-supervised pre-training to help alleviate class imbalance. 
In other domains, some methods conduct minor class data oversampling via synthetic data generation, the most notable of which is the Synthetic Minority Over-sampling Technique (SMOTE)~\citep{chawla2002smote}, which has been widely adopted, although not specifically for sepsis. In this work, we will conduct a comprehensive empirical study to explore their potential in handling our rare event problem.

%% file: S3_Dataset_icml.tex
\subsection{Cohort Extraction: Critically Ill Trauma Patients}
\label{sec:cohort}
The critically ill trauma cohort is defined using standard trauma cohort selection criteria, similar to those employed in previous studies \citep{fu2019early, li2023real}, with extraction criteria targeting critically ill patients following the guidance of \citep{stern2023defining}. The extraction process starts with the identification of patients with valid data. Specifically, valid patients are defined as those hospital admissions (HADM\_ID) that include at least one ICU stay and corresponding data in CHARTEVENTS, which serves as the primary repository for a patient’s information during their hospital stay, encompassing vital signs, laboratory values, and ventilator settings.
We then select admissions associated with traumatic injuries based on a carefully curated list of ICD-9 E-codes that broadly represent trauma-related conditions. This list excludes categories such as poisoning to ensure relevance and consistency with trauma-specific cohorts. The full list is in Appendix~\ref{ICD9_Ecode_list}.

We then refined our study cohort to include only adult patients aged between 18 and 89 years, with an admission duration of 48 hours or more. This age range is selected because the lower limit of 18 represents adult patients, consistent with standard definitions in medical research \citep{komorowski2018artificial, li2023real}. The upper limit of 89 is established because, in the MIMIC-III dataset, patients older than 89 are recorded as being 300 years old to protect their privacy \citep{johnson2016mimic}. The admission duration criterion is designed to exclude patients at low risk of sepsis, either because they died or recovered quickly. 

Finally, we included only patients with three or more days of mechanical ventilation to identify critically ill patients at higher risk of developing sepsis. Ventilation days are defined as the total number of days that each patient received mechanical ventilation during a single admission, regardless of the number of hours on each day. Although this criterion is not typically included in trauma cohort extractions, it is crucial for identifying patients with a higher likelihood of sepsis, as noted by \citep{horn2022hla, stern2023defining}. This trend is evident in the MIMIC-III dataset, where the sepsis ratio shows a rapid increase starting from patients with three days of ventilation, as in Appendix Fig.~\ref{fig:SepsisRatio_MV}.

In the MIMIC dataset, the care unit defines the type of ICU, such as the Coronary Care Unit (CCU), Trauma/Surgical ICU (TSICU), or Neonatal ICU (NICU). While many hospital admissions involve only one ICU stay, patients may transfer between these specialized ICUs based on their medical needs during a single hospital stay. Additionally, a patient can be admitted to the hospital for an extended period (e.g., a month) before being transferred to an ICU. Unlike \citep{komorowski2018artificial}, which used ICUSTAY\_ID, we selected HADM\_ID (hospital admission ID) as the instance ID for our cohort. HADM\_ID encapsulates all medical interventions and observations during a specific hospital admission, aligning with the concept of a ``patient'' in the clinical research of sepsis studies. This also aligns with how ICD-9 diagnosis codes and billing are assigned based on the entire hospital stay rather than individual ICU stays. This approach allows for comprehensive tracking of patient outcomes, providing a more complete view of the entire hospital stay. Subsequently, we gathered a refined cohort of critically ill trauma patients suitable for our study on early sepsis onset detection, comprising a total of 1,570 admissions, as summarized in Fig.~\ref{fig:FlowDiagram_Cohort} and the detailed patient characteristics are shown in Table~\ref{tab:Characteristics}.

\input{Appendices/CohortFlow}

\input{Appendices/PatientCharacteristics}

\subsection{Post-trauma Sepsis Definition}
\label{sec:Post-traumaSepsis}
This work relies on a clinical post-trauma sepsis identification defined by \citep{stern2023defining}, adapted from the Centers for Disease Control and Prevention’s adult sepsis surveillance criteria and the original Sepsis-3 guidelines. We define hospital-acquired post-traumatic sepsis within the MIMIC-III dataset as a clinically suspected infection accompanied by acute organ dysfunction, with modifications tailored to the unique challenges of classifying sepsis in our trauma cohort. This methodology involves two key steps: 1) \textbf{Pre-processing feature tables}, focusing on extracting and preprocessing relevant data from MIMIC-III to identify post-trauma sepsis cases through careful feature selection and qualified record processing based on cross-referenced multi-source data; and 2) \textbf{Post-trauma Sepsis onset criteria}, where we establish specific criteria for sepsis onset targeted at trauma patients. In the initial phase, we preprocessed pertinent data from three primary tables: blood culture, antibiotics, and a modified version of the Sequential Organ Failure Assessment (SOFA) score. The blood culture and antibiotic data are jointly analyzed to identify suspected infections, while the SOFA score is used to define the onset and quantify the severity of organ dysfunction.

\subsubsection{Pre-processing: blood culture}
For blood cultures, we apply a filter to capture relevant entries occurring at or after 72 hospital hours. We focus exclusively on blood culture criteria, in contrast to the approaches taken by AI Clinician works \citep{komorowski2018artificial} and \citep{stern2023defining}, which utilize all body tissue cultures (e.g., blood, urine, sputum). Following the methodology of \citep{rhee2019variation, reyna2020early}, we emphasize blood cultures because they are typically part of a panel of samples collected when sepsis is suspected. When clinicians suspect an infection and potential sepsis but are uncertain about the source, blood cultures are obtained alongside other targeted body fluid samples. Unlike other cultures (such as tracheal or urine cultures), which may be collected for surveillance of antibiotic-resistant organisms and might not indicate a suspected infection, blood cultures specifically aim to identify systemic infections. While this approach may result in a lower number of identified infections and sepsis cases compared to using all body fluid cultures, it is likely more specific for diagnosing systemic infections critical for sepsis identification. The filter for blood cultures at or after 72 hospital hours excludes cases of sepsis acquired before hospital admission \citep{rhee2020sepsis, stern2023defining}. 

\subsubsection{Pre-processing: antibiotic}
Pre-processing antibiotic data from the Prescriptions table in the MIMIC dataset is complicated due to the lack of standardized antibiotic labels and the disordered nature of the data, which often includes overlapping or fragmented entries and inconsistent drug name formats. Most previous work has overlooked these significant challenges, making it difficult to study and reproduce sepsis-related datasets using MIMIC. This lack of transparency leads to two major drawbacks: first, different processing methods for antibiotics can result in inaccurate sepsis identification, such as including antibiotics administered for prophylactic purposes, which may increase false positive cases; second, from a machine learning perspective, this inconsistency hampers the reproducibility of the dataset, reducing comparability when training methods are applied across different studies.

To address this gap, we propose a detailed, post-trauma sepsis-specific set of criteria for extracting antibiotic records, following the guidance of \citep{stern2023defining}, to ensure the accuracy and relevance of the data for defining sepsis onset in trauma patients. The criteria begin with identifying qualifying antibiotic drug names, selected based on cross-referencing multiple sources \citep{johnson2016mimic, komorowski2018artificial, stern2023defining} tailored specifically for the treatment of trauma-related sepsis. Further details are provided in Appendix~\ref{sec:appendix_abx}. We restrict the criteria to all intravenous (IV) antibiotics and two designated oral antibiotics—vancomycin and linezolid—while excluding prophylactic antibiotics and those administered on the first day. We also ensure that the same qualifying antibiotic is not administered within the previous day to identify new antibiotic orders accurately. To assure meaningful treatment, we also require that a qualifying antibiotic being administered for a minimum of four consecutive days or until the patient's death or discharge, without necessitating the same antibiotic throughout this period \citep{rhee2019variation, stern2023defining}. To ensure the accuracy and relevance of the data for defining hospital-acquired sepsis, the final qualified antibiotic events must meet all of the above criteria and only the starting time of the coherent antibiotic events will be used to identify the suspected infections. 

\subsubsection{Pre-processing: modified SOFA score}
Concurrently, we used a modified SOFA score targeted for trauma cohort \citep{rhee2019variation, bosch2022predictive, stern2023defining}, which excludes the urine output (uo) and Glasgow Coma Scale (GCS) variables utilized in the traditional SOFA score calculation. Excluding the Glasgow Coma Scale (GCS) from the SOFA score calculation for trauma patients is due to the confounding influence of traumatic brain injury, which can significantly alter GCS values irrespective of sepsis-related organ dysfunction. This approach aligns with historical practices \citep{minei2012changing, horn2022hla} in trauma-specific organ failure scores and helps ensure that neurological impairments caused by trauma do not skew the assessment of other organ systems. We also omitted urine output in the renal component as per a validated modification targeted on critically ill patients.

\input{Appendices/SepsisOnsetTimeline}

\subsubsection{Post-trauma sepsis onset criteria}
In the subsequent phase, we define sepsis patients based on the above pre-processed data. For each patient, we specify the following three time points to determine the onset time of sepsis \( t_{sepsis} \) as illustrated in Fig.~\ref{fig:sepsisdefination_timeline}. Concretely, $t_{infection}$ as the clinical suspicion of infection is identified as the timestamp when a blood culture is ordered within a 5-day window of antibiotic initiation for qualified antibiotic events \citep{rhee2019variation,stern2023defining}. $t_{SOFA}$ is the cccurrence of organ failure, identified by at least a 2-point increase in the modified SOFA score within a 7-day window, which includes 3 days before, the day of, and 3 days after the qualifying culture, following the guidance of \citep{reyna2020early, stern2023defining}. $t_{sepsis}$ is the onset of sepsis, identified as the earliest culture timestamp that meets the criteria for both suspicion of infection and organ failure. Since early sepsis detection focuses solely on the first onset event, this timestamp is crucial for timely intervention.


Among the 1,570 trauma admissions, we identified 729 admissions with potential infections and 535 patients with sepsis. The distribution of onset days for sepsis is depicted in Appendix Fig.~\ref{fig:TimingofSepsis_cx}. Notably, the peak for culture orders, indicating the onset of sepsis, occurs on day 5, which is consistent with clinical experience~\citep{horn2022hla}.

\subsection{Early Sepsis Detection Setup}
\label{sec:DetectionSetup}
In our dataset, we adopt a deployable daily detection setup that is closely aligned with ICU workflows. Detections are generated each morning, just before routine rounds, so that medical staff can incorporate the results into patient care decisions for the upcoming day. Beyond its clinical alignment, this setup also improves data quality, as nighttime records are generally less affected by external interventions (e.g., surgeries and diagnostic procedures).

\subsubsection{Feature extraction}
The features collected during nighttime hours (from 6:00 p.m. to 6:59 a.m. the following day) are used for early sepsis detection within the next 24 hours. We extracted seven key vital sign features: heart rate, systolic blood pressure, diastolic blood pressure, mean blood pressure, respiratory rate, temperature, and SpO2. These features are critical indicators of physiological status, which can be helpful in the early detection of sepsis. While the current feature set is sufficient for early sepsis onset detection, additional features (e.g., lab results) can be extracted and our repository will be updated and maintained accordingly in the future.

\subsubsection{Instance construction}
The raw data is first aggregated into average hourly records, converting it into a 2D time-series format with a shape of $(T, F)$, where $T = 13$ represents the 13 hourly timestamps over a night (6 p.m., 7 p.m., ..., 6 a.m.), and $F = 7$ represents the seven key vital sign features. We then filter the nights to include only those from day 2 to day 14 since the patient’s admission, focusing on the critical period for early sepsis detection. Infections occurring before day 2 are considered to have been acquired before hospital admission. Records after day 14 are excluded, as patients are classified as having chronic critical illness after this period.  

To ensure compatibility with most machine learning approaches, we also provide a standard dataset without missing values, as the ``S dataset''. Missing values were imputed using a forward and then backward filling method within the window from 7:00 a.m. (right after the end of the previous night) to 6:59 a.m. (before the end of the current night). Although backward filling is applied, this approach remains deployable as it does not rely on timestamps beyond the nighttime period. Instances that still contain missing values after this process are removed. We also include the ``N dataset'' to preserve raw data as detailed in Appendix~\ref{sec:Sdataset}.

These nighttime records were then assigned 0/1 labels based on the patient's sepsis onset time. If sepsis onset occurs within 24 hours after 6:59 a.m. of the corresponding record, it is labeled as positive; otherwise, it is labeled as negative. This means all nighttime instances of non-sepsis patients will be labeled as negative. For sepsis patients, only the night immediately preceding the sepsis onset will be labeled as positive. For example, if a nighttime instance starts at 6:00 p.m. on day $i$ and ends at 6:59 a.m. on day $(i+1)$, this instance will be labeled as positive only if the sepsis onset occurs within the window from 7:00 a.m. on day $(i+1)$ to 6:59 a.m. on day $(i+2)$. All instances of the same patient before the positive will be labeled negative, and those after will be discarded as they are not relevant to early detection. 

\input{Appendices/PartTemporalTrendsofFeatures}

Finally, our post-trauma early sepsis detection data includes 440 positive and 8,319 negative instances in the S dataset (8,759 cases across 1,522 unique patients), resulting in an imbalance ratio of 0.050 (positive/all instances). The N dataset contains 455 positive and 8,522	negative instances (8,977 cases across 1,535 unique patients), with an imbalance ratio of 0.051. The study population is smaller than the original trauma cohort (1,570 patients) due to missing data in MIMIC, and drop details are explained in Appendix~\ref{sec:Sdataset}.
It is a relatively small cohort and we will demonstrate its importance (small but high quality) in the experiment.

%% file: Appendices/CohortFlow.tex

\begin{figure}[h]
  \centering
  \includegraphics[scale=0.35]{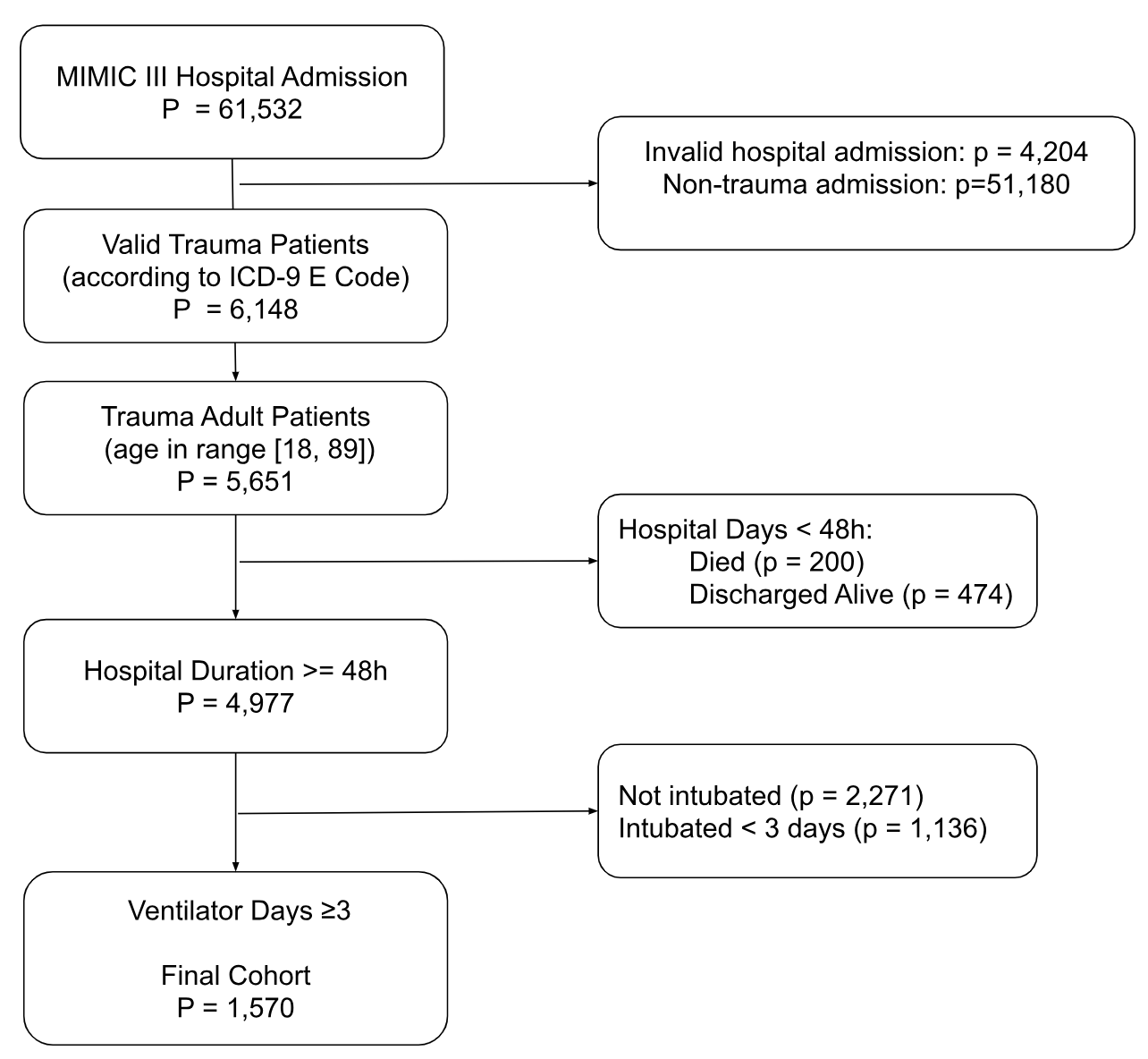}
  \caption{Data flow diagram of study inclusion and exclusion criteria.}
  \label{fig:FlowDiagram_Cohort}
\end{figure}

%% file: Appendices/PatientCharacteristics.tex
\begin{table}[!h]
\centering
\caption{Characteristics and Clinical Outcomes of Critically Ill Injured Patients Admitted to Intensive Care Unit (2012-2020)}
\label{tab:Characteristics}
\begin{center}
  \begin{small}
      \begin{tabularx}{\columnwidth}{X r}
        \toprule
        \textbf{Characteristic} & \textbf{No. (\%) (N = 1570\(^a\))} \\
        \midrule
        Age, median (IQR), y & 59 (43-75) \\
        Sex, n (\%) &  \\
        \hspace{5mm} Female & 538 (34) \\
        \hspace{5mm} Male & 1032 (66) \\
        Race, n (\%) &  \\
        \hspace{5mm} Asian & 22 (1.4) \\
        \hspace{5mm} Black & 85 (5.4) \\
        \hspace{5mm} White & 1102 (70.2) \\
        \hspace{5mm} Other & 75 (4.8) \\
        \hspace{5mm} Hispanic & 63 (4.0) \\
        \hspace{5mm} Unknown & 223 (14.2) \\
        Charlson Comorbidity Index\(^c\), median (IQR) & 12 (0-12) \\
        Injury Severity Score, median (IQR) & 16 (4-20) \\
        Body regions with an AIS \(\geq 3\), n(\%) &  \\
        \hspace{5mm} Head/Neck & 709 (45) \\
        \hspace{5mm} Chest & 395 (25) \\
        \hspace{5mm} Abdomen & 135 (9) \\
        \hspace{5mm} Lower extremity & 215 (14) \\
        LACTATE mmol/L\(^d\), median (IQR) & 3 (2-5) \\
        \hspace{5mm} Unknown, n (\%)  & 211 (13) \\
        Initial ED SBP $< 90$ mm Hg, n (\%)  & 72 (5) \\
        \textbf{Outcomes} &  \\
        \hspace{5mm} Sum of ICU days, median (IQR) & 10 (6-18) \\
        \hspace{5mm} Hospital admission days, median (IQR) & 16 (10-24) \\
        \hspace{5mm} Mechanical ventilation days, median (IQR) & 7 (4-12) \\
        \hspace{5mm} Discharged Location, n (\%) \\ 
        \hspace{10mm} Rehab/Distinct & 588 (37) \\
        \hspace{10mm} Dead & 341 (22) \\
        \hspace{10mm} Home & 289 (18) \\
        \hspace{10mm} SNF & 279 (18) \\
        \hspace{10mm} Other & 73 (5) \\
        \bottomrule
      \end{tabularx}
      
      \vspace{2mm} 

      \begin{minipage}[t]{\columnwidth}
      \raggedright
      \textbf{Abbreviations:} \\
        AIS: Abbreviated Injury Severity Score \\
        ICU: Intensive care unit \\
        LTFC: Long-term care facility \\
        SBP: Systolic blood pressure \\
        SNF: Skilled nursing facility \\
        ED: Emergency Department \\
        \(^a\) Missing values greater than 5\% are reported. \\
        \(^c\) Elixhauser-van Walraven comorbidity index.\\
        \(^d\) Highest value documented during the first 48 hours of ED admission.\\
      \end{minipage}
  \end{small}
\end{center}
\vskip -0.1in
\end{table}

%% file: Appendices/SepsisOnsetTimeline.tex

\begin{figure}[ht]
  \centering
  \includegraphics[width=0.8\textwidth]{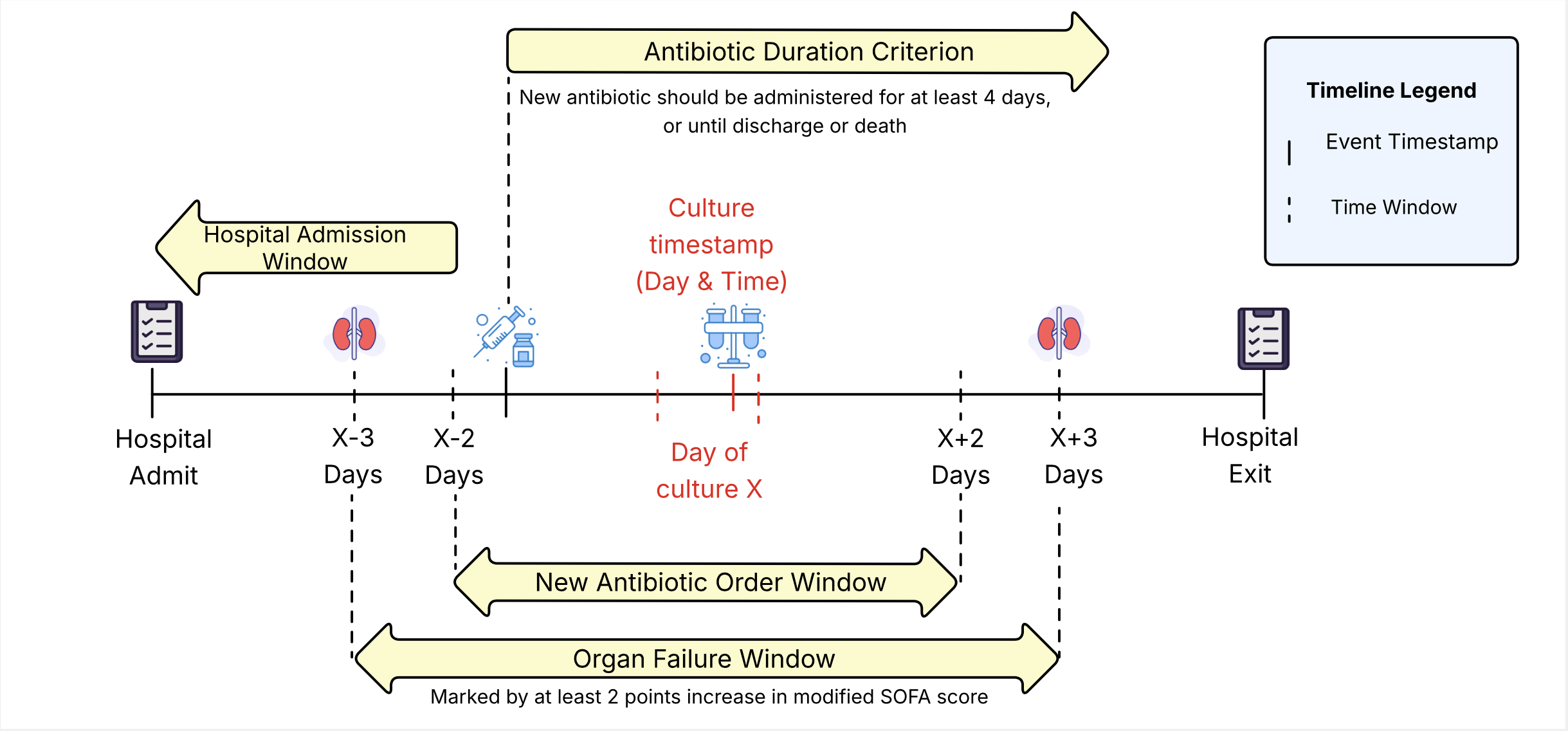}
  \caption{
    Timeline visualization of post-trauma sepsis onset assignment criteria.
  }
  \label{fig:sepsisdefination_timeline}
\end{figure}

%% file: Appendices/PartTemporalTrendsofFeatures.tex
\begin{figure}[ht]
    \vskip -0.1in
  \centering
    \includegraphics[scale=0.5]{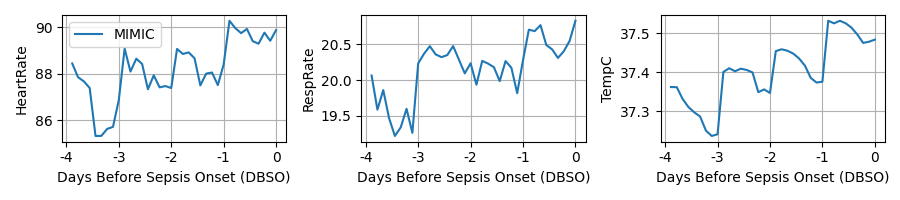}
  \caption{Subset of physiological trends in the four days before sepsis onset (x = 0 marking sepsis onset).}
  \label{fig:DBSO_trends_subset}
  \vskip -0.15in
\end{figure}

To demonstrate that these vital sign features can show early signs of sepsis, we visualize the average temporal trends of key physiological features (e.g., heart rate, temperature) in the four days leading up to sepsis onset, using Accumulated Days Before Sepsis Onset (DBSO) on the x-axis. Fig.~\ref{fig:DBSO_trends_subset} highlights noticeable changes, particularly in temperature and heart rate at x = -1, suggesting that the delta values between consecutive nights may enhance early sepsis detection. All features are visualized in Appendix~\ref{sec:DBSO_trends}.

%% file: S4_Method.tex
The rare event early detection task supported by this dataset is the early detection of sepsis onset in trauma patients using nighttime vital signs. Specifically, we formulate this problem as a binary classification problem. Let \( \{(x_{i}, x_{i}^{\Delta}), y_i\}_{i=1}^{n} \) denote the dataset, where \( {x_i} \) is a multivariate time-series of nighttime vital signs, \( x_{i}^{\Delta} \) represents the change (delta) between the current night’s data $x_i$ and its previous night's data, and \( y_i \) is the sepsis label. 

To address the severe class imbalance (i.e., rare event) problem, we propose to do reconstruction-based representation learning first to capture the general time series data distribution. Thereafter, to capture more local variance for limited sepsis instances, we can apply random masks in different timestamps for data augmentation. Specifically, as summarized in Fig.~\ref{fig:TrainingPipeline}, in Stage 1, we can adopt a Masked Autoencoder (MAE) to learn robust physiological representations from multivariate patient time series. We can explore two pre-training settings: (1) TPre: pre-training on the trauma cohort itself, and (2) GPre: pre-training on a general ICU population. When pre-training solely on the trauma cohort, we first oversample the minority class (sepsis) and then apply a masking strategy to encourage the model to learn discriminative and resilient representations.

\input{Appendices/TrainingPipeline}

In Stage 2, classification is performed on a balanced dataset obtained via oversampling, followed by masking to increase local variability. We evaluate two augmentation strategies: (1) mask-based augmentation, where the masked inputs are directly used for downstream classification, and (2) reconstruction-based augmentation, where the oversampled dataset is first reconstructed by the pre-trained autoencoder and the resulting balanced representations are used to train the classifier. In both settings, the classifier is trained on the augmented dataset to improve the robustness in detecting rare sepsis events. It should be noted that this is a general benchmark approach for demonstration while our main contribution is the new dataset. More details are in Appendix~\ref{sec:approach}.

%% file: Appendices/TrainingPipeline.tex
\begin{figure}[ht]
  \centering
    \includegraphics[width=0.8\columnwidth]{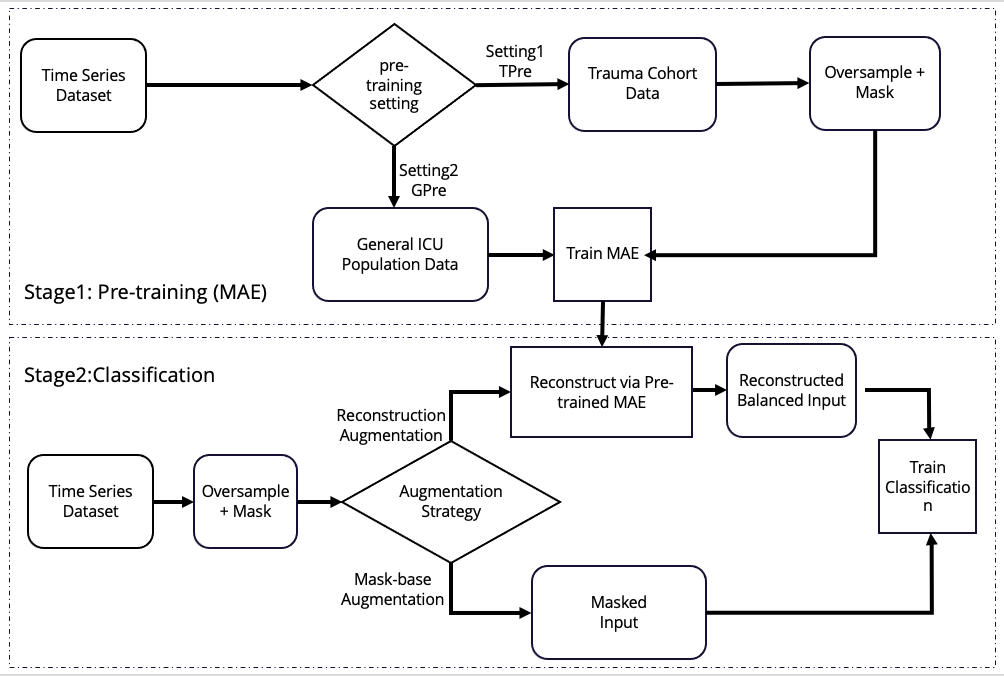}
  \caption{
    Overview of the Two-Stage Training Pipeline
  }
  \label{fig:TrainingPipeline}
  \vskip -0.1in
\end{figure}

%% file: S5_EXP_icml.tex
\begin{table*}[t]
  \caption{Baseline Performance (
        NoPre = No Pretraining; 
        TPre = Pretrained on trauma cohort; 
        GPre = Pretrained on general ICU cohort;
        * indicates training was conducted using Post-trauma Sepsis labels.)}    
  \label{tab:baseline_performance}
  \centering  
  \resizebox{\textwidth}{!}{
  \begin{tabular}{lcccccccccc}
    \toprule
     & tn & fp & fn & tp & sensitivity & specificity & precision & f1\_score & PR\_auc & ROC\_auc \\
    \midrule
    Sepsis3 - NoPre & 0 & 1320 & 0 & 82 & 1.000000 & 0.000000 & 0.058488 & 0.110512 & 0.029244 & 0.499242 \\
    Sepsis3 - GPre & 528 &	792 & 35 &	47 &	0.573171 &	0.400000 & 0.056019 & 0.102063 & 0.063371 & 0.497570 \\
    XGBoost* & 1407.00 (36.710) & 256.80 (20.800) & 75.00 (5.340) & 13.00 (3.320) & 0.15 (0.04) & 0.85 (0.01) & 0.05 (0.01) & 0.07 (0.02) & 0.05 (0.01) & 0.53 (0.03) \\
    LightGBM* & 1126.00 (120.84) & 537.80 (109.36) & 54.80 (9.550) & 33.20 (7.160) & 0.38 (0.09) & 0.68 (0.07) & 0.06 (0.00) & 0.10 (0.01) & 0.06 (0.00) & 0.55 (0.02) \\
    GRU-TCNN - NoPre* & 1018.40 (112.54) & 327.80 (95.050) & 58.80 (8.170) & 23.20 (7.010) & 0.28 (0.09) & 0.76 (0.07) & 0.07 (0.02) & 0.11 (0.02) & 0.07 (0.01) & 0.55 (0.05) \\
    GRU-TCNN - TPre* & 665.200 (66.980) & 681.00 (83.390) & 32.80 (11.30) & 49.20 (14.65) & 0.59 (0.16) & 0.49 (0.05) & 0.07 (0.01) & 0.12 (0.03) & 0.07 (0.01) & 0.58 (0.04) \\
    GRU-TCNN - GPre* & 723.200 (62.810) & 623.00 (60.110) & 31.00 (3.000) & 51.00 (6.780) & 0.62 (0.05) & 0.54 (0.04) & 0.08 (0.01) & 0.14 (0.02) & 0.08 (0.01) & 0.60 (0.03) \\
    \bottomrule
    \end{tabular}
  }
  \vskip -0.1in
\end{table*}

\begin{table*}[t]
\caption{Augmentation performance reported as mean (std).}
\label{tab:augmentation_performance}
\centering
\scriptsize  
\resizebox{\textwidth}{!}{
    \begin{tabular}{lllcccccccccc}
    \toprule
     & pretrain & freeze\_encoders  & tn & fp & fn & tp & sensitivity & specificity & precision & f1\_score & PR\_auc & ROC\_auc \\
    \midrule
    
    SMOTE        & NO          & FALSE & 993.60 (110.65) & 352.60 (99.640) & 52.80 (8.070) & 29.20 (9.420) & 0.35 (0.10) & 0.74 (0.07) & 0.08 (0.01) & 0.12 (0.02) & 0.07 (0.01) & 0.57 (0.01) \\
    Time Warp    & NO          & FALSE & 564.40 (303.10) & 781.80 (318.49) & 28.00 (21.52) & 54.00 (20.41) & 0.66 (0.26) & 0.42 (0.23) & 0.07 (0.01) & 0.12 (0.02) & 0.06 (0.01) & 0.55 (0.03) \\
    Noise        & NO          & FALSE & 337.20 (177.11) & 1009.00 (176.41) & 14.80 (9.680) & 67.20 (5.930) & 0.82 (0.11) & 0.25 (0.13) & 0.06 (0.01) & 0.12 (0.01) & 0.07 (0.02) & 0.56 (0.05) \\
    SMOTE        & general icu & TRUE  & 827.60 (56.730)  & 518.60 (73.490)  & 42.20 (5.260) & 39.80 (8.500) & 0.48 (0.09) & 0.62 (0.05) & 0.07 (0.01) & 0.12 (0.02) & 0.07 (0.01) & 0.57 (0.04) \\
    Time Warp    & general icu & TRUE  & 735.20 (112.43) & 611.00 (136.21) & 35.20 (7.360) & 46.80 (7.050) & 0.57 (0.08) & 0.55 (0.09) & 0.07 (0.01) & 0.13 (0.02) & 0.08 (0.02) & 0.58 (0.03) \\
    Noise        & general icu & TRUE  & 833.60 (136.33) & 512.60 (107.66) & 41.40 (10.19) & 40.60 (11.10) & 0.49 (0.13) & 0.62 (0.09) & 0.07 (0.01) & 0.13 (0.02) & 0.07 (0.01) & 0.58 (0.04) \\
    Mask         & general icu & TRUE  & 717.60 (92.740)  & 628.60 (73.130)  & 32.20 (2.950) & 49.80 (4.660) & 0.61 (0.04) & 0.53 (0.06) & 0.07 (0.01) & 0.13 (0.01) & 0.07 (0.01) & 0.59 (0.03) \\
                 & general icu & FALSE & 518.60 (427.31) & 827.60 (400.44) & 25.20 (25.03) & 56.80 (22.97) & 0.70 (0.29) & 0.38 (0.31) & 0.07 (0.01) & 0.12 (0.01) & 0.07 (0.02) & 0.56 (0.07) \\
    Reconstract  & general icu & TRUE  & 532.60 (729.63) & 813.60 (743.10) & 33.20 (45.48) & 48.80 (44.75) & 0.60 (0.55) & 0.40 (0.55) & 0.03 (0.03) & 0.06 (0.06) & 0.06 (0.01) & 0.52 (0.03) \\
                 & general icu & FALSE & 679.80 (128.29) & 666.40 (119.66) & 31.20 (6.020) & 50.80 (3.830) & 0.62 (0.06) & 0.50 (0.09) & 0.07 (0.01) & 0.13 (0.02) & 0.07 (0.01) & 0.58 (0.03) \\
    
    \bottomrule
\end{tabular}
}
\end{table*}

\begin{table*}[t!]
\caption{Performance of different model architectures reported as mean (std).}
\label{tab:model_architectures_performance}
\centering
\scriptsize
\resizebox{\textwidth}{!}{
    \begin{tabular}{lcccccccccc}
    \toprule
    Model & tn & fp & fn & tp & sensitivity & specificity & precision & f1\_score & PR\_auc & ROC\_auc \\
    \midrule
    1encoder  & 844.20 (129.66) & 819.60 (118.43) & 32.80 (9.230) & 55.20 (8.870) & 0.63 (0.10) & 0.51 (0.08) & 0.06 (0.01) & 0.11 (0.01) & 0.06 (0.01) & 0.59 (0.02) \\
    TCNNonly  & 617.20 (99.470)  & 729.00 (109.88) & 25.40 (5.860) & 56.60 (8.740) & 0.69 (0.08) & 0.46 (0.08) & 0.07 (0.01) & 0.13 (0.02) & 0.08 (0.02) & 0.60 (0.03) \\
    GRUonly   & 1098.20 (190.05) & 248.00 (214.97) & 62.60 (17.40) & 19.40 (15.19) & 0.24 (0.19) & 0.82 (0.16) & 0.10 (0.06) & 0.10 (0.04) & 0.07 (0.01) & 0.56 (0.05) \\
    \bottomrule
    \end{tabular}
}
\end{table*}

In Table~\ref{tab:baseline_performance} for baseline comparison, Sepsis3-NoPre and Sepsis3-GPre report classification results obtained by training on the publicly available Sepsis-3 labels derived from the general ICU cohort and evaluating on our trauma cohort with post-trauma sepsis labels. Because the general ICU dataset is sufficiently large, we did not apply 5-fold cross-validation for these two settings. For the rest of the experiments (XGBoost, LightGBM, GRU variants), models were trained/fine-tuned on the trauma-only dataset and evaluated using 5-fold cross-validation. To address severe class imbalance, all classifiers were trained on a balanced dataset obtained by randomly oversampling the minority (sepsis) class until it matches the size of the majority class. 

First, we can observe that baselines without pre-training show very biased performance towards one of these two classes, indicating that simple oversampling cannot alleviate the severe class imbalance problem well. On the other hand, all baselines with pre-training provide better balanced results between the major and minor classes, which demonstrates the importance of pre-training for severe class imbalance. Then, we can observe that fine-tuning using a significantly smaller amount of reliable post-traumatic sepsis labels compared to the large amount of general ICU sepsis labels, provides significantly better performance, which demonstrates the contribution of this new dataset. Finally, it is expected that pre-training using larger general ICU population can be more helpful.

In Table~\ref{tab:augmentation_performance}, we investigate how oversampling combined with different augmentation strategies, as well as the use of pretrained weights, affects rare event detection performance. Based on the observations from Table~\ref{tab:baseline_performance}, where pretraining on the general ICU cohort yields more consistent results than pretraining on the trauma cohort, we restrict the comparison here to models trained without pretraining and models pretrained on the general ICU cohort. For augmentation, we include the widely used SMOTE baseline, and for time-series augmentation we evaluate two common strategies: time warping and additive noise. In addition, we consider a task-specific augmentation derived from the Masked Autoencoder, which introduces variability through reconstruction-based masking. All experiments in this table are conducted with 5-fold cross-validation on the trauma-only dataset, with minority class samples oversampled to match the majority class size. Again, the importance of pre-training for rare event detection can be easily observed, while the masking and reconstruction-based augmentation provides better performance by covering more data variance.

In Table~\ref{tab:model_architectures_performance}, we conducted an ablation study to evaluate the contribution of different encoder design choices and the effectiveness of using a single versus a dual encoder. The dual encoder setup leverages both current-night observations and delta values, while the single encoder setup only processes the current-night input stream. All experiments in this table are initialized with pretrained weights from the general ICU cohort to ensure consistent feature representations. For the downstream classification task, we apply random oversampling to balance the classes, but no additional data augmentation is used. This design allows us to isolate the effects of encoder architecture and input configuration on performance. We can observe that after pre-training on general ICU data, the contribution of delta values between two nights is subtle, while the combination of TCNN and GRU as the backbone is important to reduce the bias between the minor and major classes. 

We also explored VAE-based pretraining, but did not observe consistent advantage over MAE in representation quality or downstream detection performance as illustrated in Appendix~\ref{VAE_performence}. 
In summary, we can observe great future opportunities to advance the rare event early detection performance using this dataset.

%% file: Appendix_ICML.tex
\section{Cohort Details}
\label{ICD9_Ecode_list}
\input{Appendices/ICD9_code_full_list}

\subsection{Sepsis Ratio and Ventilation Days}
\input{Appendices/VentilationDayFigure}

\section{Sepsis Identification Details}
\subsection{Antibiotic List}
\label{sec:appendix_abx}
This appendix provides a complete list of antibiotics used in the study. The antibiotics were selected based on established guidelines and prior research \citep{stern2023defining, johnson2016mimic, komorowski2018artificial}, ensuring relevance to trauma-related sepsis. Table~\ref{tab:AbxList} details the included antibiotics, their administration routes, and whether they are classified as prophylactic.

\begin{longtable}{@{}llll@{}}
\caption{List of Drugs}\\
\label{tab:AbxList}
gsn & drug & route & isProphylactic \\
\midrule
\endhead
\bottomrule
\multicolumn{4}{r}{{Continued on next page}} \\
\endfoot
\bottomrule
\endlastfoot
8854 & Penicillin G Potassium & IV & 0 \\
8920 & Ampicillin-Sulbactam & IV & 1 \\
8921 & Ampicillin-Sulbactam & IV & 1 \\
8921 & Ampicillin-Sulbactam & IV & 1 \\
8932 & Ampicillin Sodium & IV & 0 \\
8935 & Ampicillin Sodium & IV & 0 \\
8937 & Ampicillin Sodium & IV & 0 \\
8965 & Nafcillin & IV & 0 \\
9143 & Cefuroxime Sodium & IV & 0 \\
9144 & Cefuroxime Sodium & IV & 0 \\
9156 & CefTRIAXone & IV & 0 \\
9156 & Ceftriaxone & IV & 0 \\
9157 & CefTRIAXone & IV & 0 \\
9157 & Ceftriaxone & IV & 0 \\
9162 & CeftriaXONE & IV & 0 \\
9165 & CeftriaXONE & IV & 0 \\
9171 & CefTAZidime & IV & 0 \\
9172 & CefTAZidime & IV & 0 \\
9172 & CeftazIDIME & IV & 0 \\
9181 & Cefotetan & IV & 0 \\
9181 & Cefotetan & IV & 0 \\
9221 & Doxycycline Hyclate & IV & 0 \\
9251 & Erythromycin & IV & 1 \\
9251 & Erythromycin Lactobionate & IV & 1 \\
9252 & Erythromycin & IV & 1 \\
9289 & Gentamicin & IV & 0 \\
9291 & Gentamicin & IV & 0 \\
9294 & Gentamicin & IV & 0 \\
9299 & Gentamicin & IV & 0 \\
9299 & Gentamicin Sulfate & IV & 0 \\
9312 & Amikacin & IV & 0 \\
9328 & Vancomycin & IV & 0 \\
9329 & Vancomycin & IV & 0 \\
9329 & Vancomycin HCl & IV & 0 \\
9329 & Vancomycin Oral Liquid & PO & 0 \\
9329 & Vancomycin Oral Liquid & PO/NG & 0 \\
9330 & Vancomycin & IV & 0 \\
9331 & Vancomycin & IV & 0 \\
9331 & Vancomycin Enema & IV & 0 \\
9331 & Vancomycin HCl & IV & 0 \\
9344 & Clindamycin & IV & 0 \\
9344 & Clindamycin Phosphate & IV & 0 \\
9361 & Aztreonam & IV & 0 \\
9362 & Aztreonam & IV & 0 \\
9365 & Imipenem-Cilastatin & IV & 0 \\
9393 & Sulfameth/Trimethoprim & IV & 0 \\
9393 & Sulfamethoxazole-Trimethoprim & IV & 0 \\
9525 & Amphotericin B & IV & 0 \\
9588 & MetRONIDAZOLE (FLagyl) & IV & 0 \\
9588 & Metronidazole & IV & 0 \\
9592 & Metronidazole & IV & 0 \\
13052 & Clindamycin & IV & 0 \\
13053 & Clindamycin & IV & 0 \\
13645 & Rifampin & IV & 0 \\
14196 & Aztreonam & IV & 0 \\
14197 & Aztreonam & IV & 0 \\
15327 & Gentamicin & IV & 0 \\
15355 & Nafcillin & IV & 0 \\
15538 & CeftAZIDime & IV & 0 \\
15538 & CeftazIDIME & IV & 0 \\
15538 & Ceftazidime & IV & 0 \\
15539 & CeftAZIDime & IV & 0 \\
15539 & CeftazIDIME & IV & 0 \\
15539 & Ceftazidime & IV & 0 \\
15920 & Ciprofloxacin IV & IV & 0 \\
15921 & Ciprofloxacin & IV & 0 \\
15921 & Ciprofloxacin IV & IV & 0 \\
15932 & Penicillin G Potassium & IV & 0 \\
15933 & Penicillin G Potassium & IV & 0 \\
15934 & Penicillin G Potassium & IV & 0 \\
21185 & Piperacillin-Tazobactam & IV & 0 \\
21185 & Piperacillin-Tazobactam Na & IV & 0 \\
21187 & Piperacillin-Tazobactam & IV & 0 \\
21187 & Piperacillin-Tazobactam Na & IV & 0 \\
21701 & Cefotetan & IV & 0 \\
21702 & Cefotetan & IV & 0 \\
24094 & CefePIME & IV & 0 \\
24095 & CefePIME & IV & 0 \\
26488 & Meropenem & IV & 0 \\
26489 & Meropenem & IV & 0 \\
27468 & CefePIME & IV & 0 \\
27468 & Cefepime & IV & 0 \\
27470 & Cefepime & IV & 0 \\
29925 & Levofloxacin & IV & 0 \\
29927 & Levofloxacin & IV & 0 \\
29928 & Levofloxacin & IV & 0 \\
29929 & Levofloxacin & IV & 0 \\
31452 & Azithromycin & IV & 0 \\
31535 & Nafcillin & IV & 0 \\
40819 & Piperacillin-Tazobactam & IV & 0 \\
40819 & Piperacillin-Tazobactam & IV & 0 \\
40819 & Piperacillin-Tazobactam Na & IV & 0 \\
40819 & Piperacillin-Tazobactam Na & IV & 0 \\
43952 & Vancomycin & IV & 0 \\
43952 & Vancomycin HCl & IV & 0 \\
45131 & Linezolid & PO/NG & 0 \\
45131 & Linezolid & PO & 0 \\
45134 & Linezolid & IV & 0 \\
45134 & Linezolid & PO/NG & 0 \\
46770 & Levofloxacin & IV & 0 \\
46771 & Levofloxacin & IV & 0 \\
57824 & Ciprofloxacin & IV & 0 \\
57825 & Ciprofloxacin & IV & 0 \\
57825 & Ciprofloxacin IV & IV & 0 \\
59424 & CefTRIAXone & IV & 0 \\
59424 & CeftriaXONE & IV & 0 \\
59425 & CefTRIAXone & IV & 0 \\
59425 & CeftriaXONE & IV & 0 \\
59747 & CefTAZidime & IV & 0 \\
59747 & CeftazIDIME & IV & 0 \\
\end{longtable}


\subsection{Sepsis Onset Timing Distribution}
This histogram provides a visual representation of the timing of sepsis onset relative to hospital admission and antibiotic initiation.
\input{Appendices/TimingDistribution}

\subsection{Temporal Trends of Physiological Features Before Sepsis Onset}
\label{sec:DBSO_trends}
\input{Appendices/FullTemporalTrendsofFeatures}

\section{Detection Setup}
\label{sec:Sdataset}
Since handling irregular time-series data and imputing missing values are widely discussed topics in time-series analysis, we also include the ``N dataset'' (which retains NaN values) to preserve the raw form of the data without introducing potential imputation bias. Notably, the ``S dataset'' is a subset of the ``N dataset'' since some instances cannot be fully filled using our method and window constraints. Additionally, if no recorded data exists for a specific patient during a given night in the ``N dataset'', that instance cannot appear in the ``S dataset''. This ensures that we do not introduce artificially generated values that would be identical across all timestamps, which can mislead the model.
In the N dataset, we excluded several instances for specific reasons: 9 patients had data limited to nightly observations after day 14 (6 of them were sepsis patients), which falls outside the critical period for early sepsis detection. Additionally, we removed 26 sepsis patients who only had data available after their sepsis onset and 48 patients for whom we could not locate positive instances. These positive instances were often located outside ICU stays, occurring before, between, or after ICU stays. Since vital signs in the MIMIC dataset are only documented during ICU stays, this complicates the tracking of early sepsis signs (i.e., the positive instance). Furthermore, excluding data from outside the ICU is justified, as these patients may have transitioned between units or been hospitalized for extended periods prior to ICU admission, potentially altering the reasons for their sepsis. Ultimately, our final dataset for the N subset includes 455 positive instances and 8,522 negative instances, totaling 8,977 cases across 1,535 unique patients.

\subsection{Cross-Validation Data Splitting Strategy}
\label{sec: 5fold_CV}

The data splitting process aims to prevent data leakage and ensure a balanced representation of sepsis and non-sepsis patients across subsets. In the MIMIC dataset, each individual is assigned a unique SUBJECT\_ID, which may encompass one or multiple hospital admissions. The time gaps between these admissions can vary significantly, as patients may be admitted for different reasons. Although the hospital admission ID (HADM\_ID) aligns more closely with the clinical concept of a patient, we perform the split based on SUBJECT\_ID. Specifically, we define a sepsis individual as one for whom at least one hospital admission is identified as sepsis, and then execute a stratified split based on SUBJECT\_ID to ensure that all records for each individual remain within the same subset.

Given the limited data size, we employ stratified 5-fold cross-validation to ensure balanced label distribution across folds. In each iteration, one fold is used for testing, while the remaining folds are used for training. Since the dataset is small, the same fold serves both as the validation set during training and as the test set for final evaluation. The key difference is that oversampling is applied to the validation set during training to address class imbalance, whereas the test set remains untouched to preserve the original data distribution. Across all folds, the class imbalance ratio (positive/total) ranges around 0.05. For further details, refer to Table~\ref{tab:5fold_CV_S} for ``S Dataset'' and Table~\ref{tab:5fold_CV_N} for ``N Dataset''.

\begin{table}[h]
\caption{Distribution of the S Dataset across folds.}
\label{tab:5fold_CV_S}
\centering
\scriptsize
\begin{tabular}{lccccc}
\toprule
Fold & Total Instances & Positive Instances & Negative Instances & Imbalance Ratio \\
\midrule
0     & 1735 & 88  & 1647 & 0.0507 \\
1     & 1763 & 87  & 1676 & 0.0493 \\
2     & 1797 & 93  & 1704 & 0.0518 \\
3     & 1711 & 90  & 1621 & 0.0526 \\
4     & 1753 & 82  & 1671 & 0.0468 \\
\midrule
Total & 8759 & 440 & 8319 & 0.0502 \\
\bottomrule
\end{tabular}
\end{table}

\begin{table}[h]
\caption{Distribution of the N Dataset across folds.}
\label{tab:5fold_CV_N}
\centering
\scriptsize
\begin{tabular}{lccccc}
\toprule
Fold & Total Instances & Positive Instances & Negative Instances & Imbalance Ratio \\
\midrule
0     & 1788 & 92  & 1696 & 0.0515 \\
1     & 1813 & 90  & 1723 & 0.0496 \\
2     & 1825 & 94  & 1731 & 0.0515 \\
3     & 1740 & 92  & 1648 & 0.0529 \\
4     & 1811 & 87  & 1724 & 0.0480 \\
\midrule
Total & 8977 & 455 & 8522 & 0.0507 \\
\bottomrule
\end{tabular}
\end{table}

\section{The Benchmark Approach}
\label{sec:approach}
\subsection{Model Architecture}
\textbf{Encoder} We use a time-series encoder that combines a Bidirectional GRU for modeling temporal dependencies with 1D convolutional layers for capturing local feature correlations.
To mitigate overfitting, kernel regularization, batch normalization, and dropout are applied.
The encoder outputs a fixed-length representation via a pooling layer for downstream tasks.

\textbf{Autoencoder}  We adopt a multi-modal autoencoder to jointly model current-night and delta inputs using two separate encoders, one for each time-series modality. The resulting embeddings are concatenated into a unified latent representation and reconstructed by a shared decoder. This reconstruction-based pretraining is used to learn data representations in a self-supervised manner and provide initialization for downstream classification.

\textbf{Classifier} The classifier takes the encoded feature representation from the pre-trained encoder and applies a fully connected classification head to predict the sepsis outcome.

\subsection{Training Pipeline}

\textbf{Stage 1: masked autoencoder (MAE) for feature learning} To further emphasize masked regions, we adopt a time-series targeted masking scheme of \citep{zerveas2021transformer}, which generates contiguous segments of masked values for each variable with lengths drawn from a geometric distribution, and apply a weighted Mean Absolute Error loss, where a hyperparameter \( \lambda \) controls the relative weighting between masked and unmasked values.

\section{Performance of VAE-based}
\label{VAE_performence}
\input{Appendices/VAE}

%% file: Appendices/ICD9_code_full_list.tex
We selected admissions associated with traumatic injuries based on a carefully curated list of ICD-9 E-codes that broadly represent trauma-related conditions, that is, E8000-E8480, E8800-E9057, E9060-9259, E9270-E9289, E9507, E9520, E9521, E9520-E9589, E9600, E9610- E9689, E9680-E9689, E9700-E9760, E9780-E9799, E9806, E9830-E9886, E9888, E9889, E9900-E9961, and E9968-E9989. We explicitly exclude poisoning-related codes (e.g., E850–E854) to ensure compatibility with trauma-specific datasets that may be developed in future studies. These codes are excluded because patients with poisoning are less likely to develop sepsis. 

%% file: Appendices/VentilationDayFigure.tex
\begin{figure}[h]
  \centering
  \includegraphics[scale=0.3]{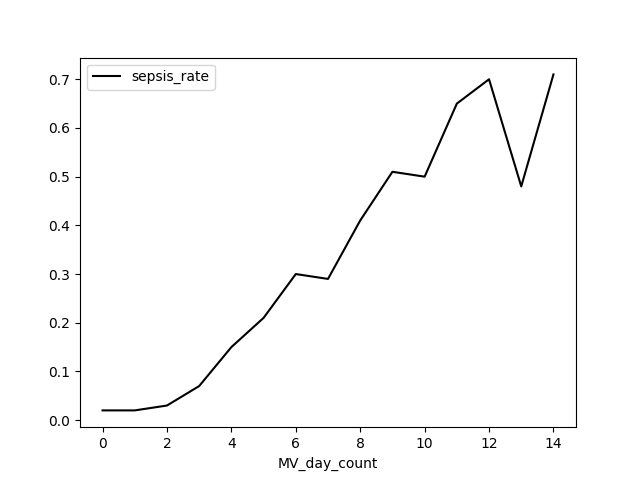}
  \caption{Sepsis ratio vs. number of ventilation days.}
  \label{fig:SepsisRatio_MV}
\end{figure}

Fig.~\ref{fig:SepsisRatio_MV} illustrates the relationship between the sepsis ratio and the number of ventilation days in the MIMIC-III dataset. The x-axis represents the total number of days a patient received mechanical ventilation during a single admission, while the y-axis denotes the corresponding sepsis ratio. The figure highlights a rapid increase in the sepsis ratio starting from patients with three or more days of mechanical ventilation, supporting the rationale for using this threshold to identify critically ill trauma patients at higher risk of developing sepsis.

%% file: Appendices/TimingDistribution.tex

\begin{figure}[ht]
  \centering
  \includegraphics[width=\columnwidth]{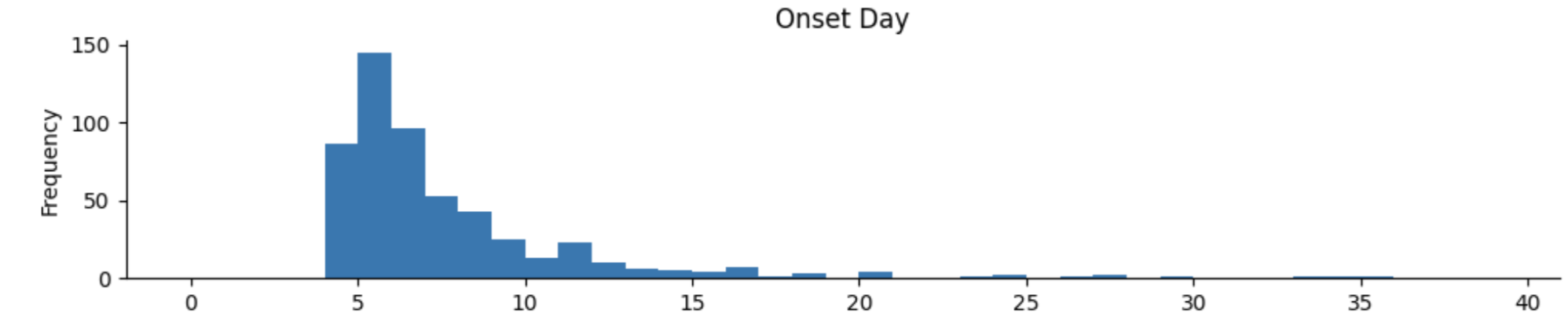}
  \caption{
    Sepsis onset day distribution after admission
  }
  \label{fig:TimingofSepsis_cx}
  \vskip -0.1in
\end{figure}

%% file: Appendices/FullTemporalTrendsofFeatures.tex
\begin{figure}[ht]
  \centering
  \includegraphics[width=\columnwidth]{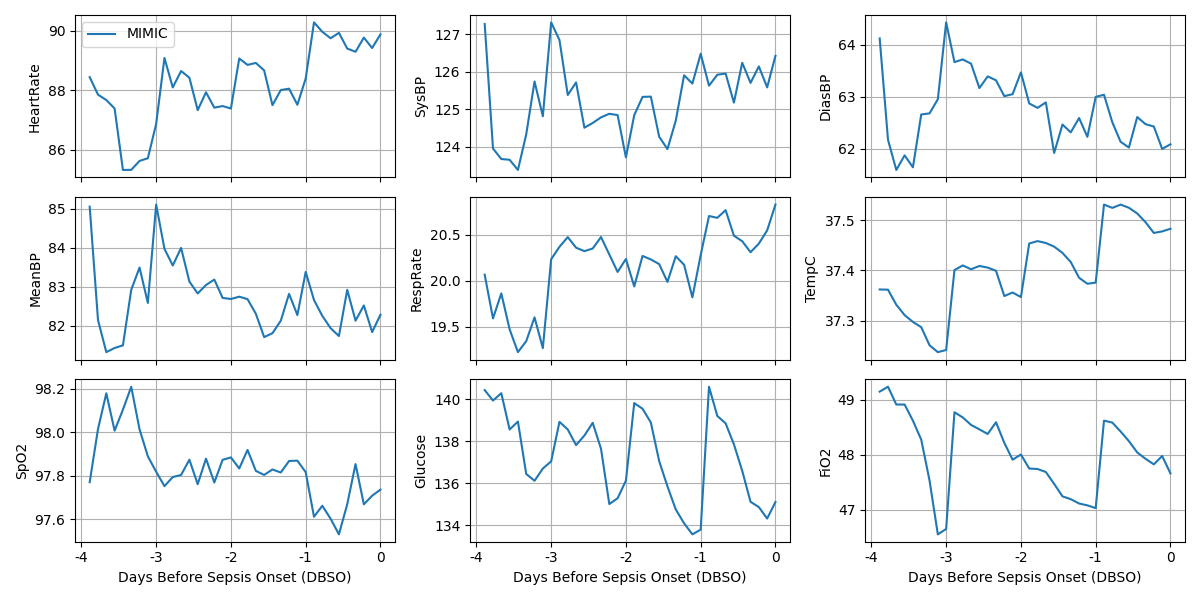}
  \caption{Temporal trends of key physiological features in the four days leading up to sepsis onset.}
  \label{fig:DBSO_trends}
  \vskip -0.1in
\end{figure}

To visualize early signs of sepsis, we plot the temporal trends of key physiological features (e.g., heart rate, blood pressure) in the four days leading up to sepsis onset. The x-axis represents time relative to sepsis onset, measured in Accumulated Days Before Sepsis Onset (DBSO), while the y-axis shows the average physiological measurements across all sepsis patients. In this representation, x = -4 corresponds to four days before sepsis onset, and x = 0 marks the onset of sepsis. The range from x = -1 to 0 represents positive nighttime samples. As shown in Fig.~\ref{fig:DBSO_trends}, certain features, particularly temperature (Temp) and heart rate (HR), exhibit noticeable changes at the beginning of the positive night (x = -1). Based on this observation and individual patient physiological trends, we hypothesize that the delta values—the difference between a patient’s current night and their previous night—can provide valuable information for early sepsis detection. Therefore, we compute these delta values as an auxiliary information to capture early signs of sepsis.

%% file: Appendices/VAE.tex

\input{Appendices/PCA_plot}
In addition to our MAE-based pretraining framework, we include an auxiliary set of experiments using a variational autoencoder (VAE) for representation learning, in order to examine the applicability of anomaly-detection-based assumptions in our setting. However, when using VAE pretraining, the resulting PCA embeddings do not exhibit the clear class separability as reported in \citep{9629559}. In our case, as shown in Fig.~\ref{fig:pca-encoder-comparison},  substantial overlap can be observed between classes, both for Sepsis-3 vs. non–Sepsis-3 and post-traumatic sepsis vs. non–post-traumatic sepsis tasks. These observations suggest that the assumption that anomalous samples form a distinct distribution does not hold in our daily sepsis detection setting.

Furthermore, Table~\ref{tab:vae_evaluating} shows VAE-pretrained representations yield downstream detection performance comparable to that of MAE-based pretraining, but do not demonstrate the clear advantage of VAE-based representations for classification reported in prior septic shock onset studies. This further highlights the difference between patient-level septic shock prediction and early-stage daily post-trauma sepsis detection in trauma cohorts, where disease signals are weaker and more heterogeneous.

\begin{table*}[t]
\caption{Performance using VAE-Based Representations. Results are reported as mean (standard deviation).}
\label{tab:vae_evaluating}
\centering
\scriptsize
\resizebox{\textwidth}{!}{
    \begin{tabular}{llcccccccccc}
    \toprule
     Augmrntation & freeze\_encoders 
     & tn & fp & fn & tp 
     & sensitivity & specificity & precision 
     & f1\_score & PR\_auc & ROC\_auc \\
    \midrule
    
    Reconstract & TRUE
    & 757.60 (60.55) & 588.60 (54.20) & 33.60 (4.62) & 48.40 (5.03)
    & 0.59 (0.05) & 0.56 (0.04) & 0.08 (0.01)
    & 0.14 (0.02) & 0.08 (0.01) & 0.61 (0.02) \\
    
                & FALSE
    & 763.80 (52.05) & 582.40 (75.61) & 33.80 (6.10) & 48.20 (5.36)
    & 0.59 (0.07) & 0.57 (0.05) & 0.08 (0.01)
    & 0.14 (0.01) & 0.08 (0.00) & 0.60 (0.02) \\
    
    Mask & TRUE
    & 703.20 (61.46) & 643.00 (76.23) & 28.60 (7.77) & 53.40 (5.55)
    & 0.65 (0.08) & 0.52 (0.05) & 0.08 (0.01)
    & 0.14 (0.01) & 0.08 (0.00) & 0.60 (0.02) \\
    
        & FALSE
    & 787.60 (45.82) & 558.60 (62.03) & 32.20 (1.64) & 49.80 (5.07)
    & 0.61 (0.03) & 0.59 (0.04) & 0.08 (0.01)
    & 0.14 (0.02) & 0.08 (0.01) & 0.60 (0.02) \\
    
    \bottomrule
    \end{tabular}
}
\end{table*}

%% file: Appendices/PCA_plot.tex
\begin{figure}[t]
  \centering

  \begin{subfigure}{\columnwidth}
    \centering
    \includegraphics[width=\columnwidth]{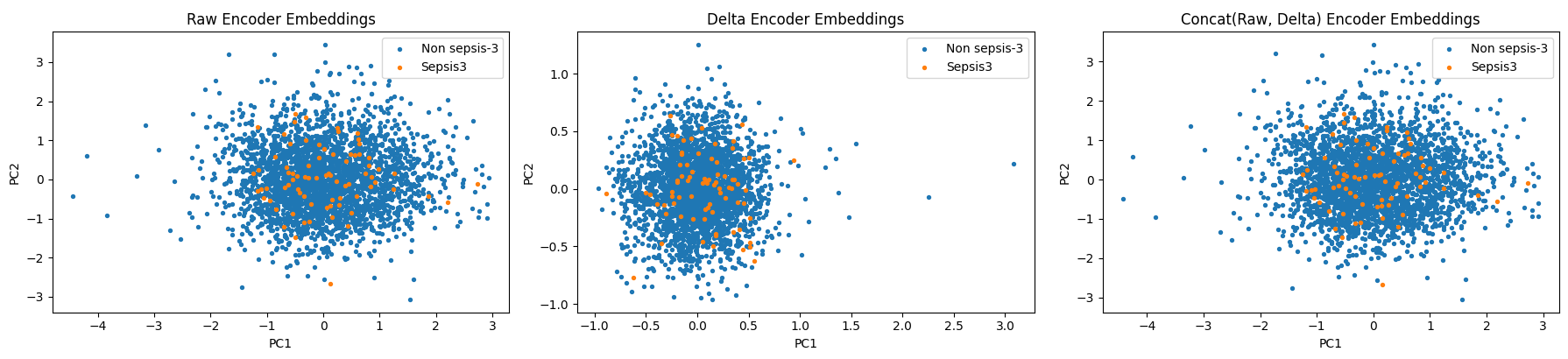}
  \end{subfigure}

  \vspace{0.15in}

  \begin{subfigure}{\columnwidth}
    \centering
    \includegraphics[width=\columnwidth]{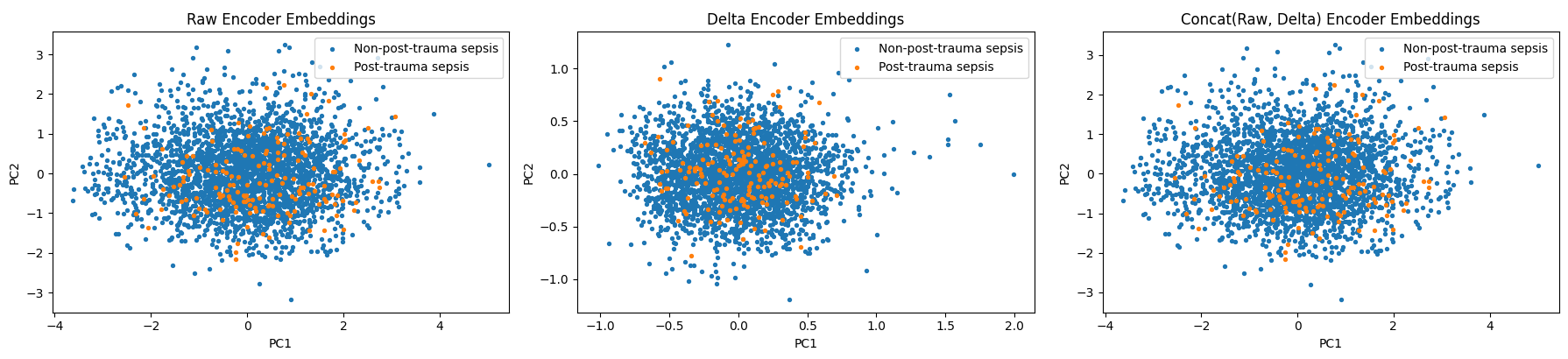}
  \end{subfigure}

  \caption{
    PCA visualization of encoder embeddings under different labeling schemes.
    The top panel shows embeddings from encoders pretrained on the general ICU cohort,
    while the bottom panel shows embeddings from encoders pretrained on the trauma cohort.
    Each panel presents PCA projections of raw, delta, and concatenated encoder embeddings.
  }
  \label{fig:pca-encoder-comparison}
  \vskip -0.1in
\end{figure}

%% file: uwSepsis.bib
@article{singer2016third,
  title={The third international consensus definitions for sepsis and septic shock (Sepsis-3)},
  author={Singer, Mervyn and Deutschman, Clifford S and Seymour, Christopher Warren and Shankar-Hari, Manu and Annane, Djillali and Bauer, Michael and Bellomo, Rinaldo and Bernard, Gordon R and Chiche, Jean-Daniel and Coopersmith, Craig M and others},
  journal={Jama},
  volume={315},
  number={8},
  pages={801--810},
  year={2016},
  publisher={American Medical Association}
}

@article{fleischmann2016assessment,
  title={Assessment of global incidence and mortality of hospital-treated sepsis. Current estimates and limitations},
  author={Fleischmann, Carolin and Scherag, Andr{\'e} and Adhikari, Neill KJ and Hartog, Christiane S and Tsaganos, Thomas and Schlattmann, Peter and Angus, Derek C and Reinhart, Konrad},
  journal={American journal of respiratory and critical care medicine},
  volume={193},
  number={3},
  pages={259--272},
  year={2016},
  publisher={American Thoracic Society}
}

@article{rudd2020global,
  title={Global, regional, and national sepsis incidence and mortality, 1990--2017: analysis for the Global Burden of Disease Study},
  author={Rudd, Kristina E and Johnson, Sarah Charlotte and Agesa, Kareha M and Shackelford, Katya Anne and Tsoi, Derrick and Kievlan, Daniel Rhodes and Colombara, Danny V and Ikuta, Kevin S and Kissoon, Niranjan and Finfer, Simon and others},
  journal={The Lancet},
  volume={395},
  number={10219},
  pages={200--211},
  year={2020},
  publisher={Elsevier}
}

@article{adams2022prospective,
  title={Prospective, multi-site study of patient outcomes after implementation of the TREWS machine learning-based early warning system for sepsis},
  author={Adams, Roy and Henry, Katharine E and Sridharan, Anirudh and Soleimani, Hossein and Zhan, Andong and Rawat, Nishi and Johnson, Lauren and Hager, David N and Cosgrove, Sara E and Markowski, Andrew and others},
  journal={Nature medicine},
  volume={28},
  number={7},
  pages={1455--1460},
  year={2022},
  publisher={Nature Publishing Group US New York}
}

@article{komorowski2018artificial,
  title={The artificial intelligence clinician learns optimal treatment strategies for sepsis in intensive care},
  author={Komorowski, Matthieu and Celi, Leo A and Badawi, Omar and Gordon, Anthony C and Faisal, A Aldo},
  journal={Nature medicine},
  volume={24},
  number={11},
  pages={1716--1720},
  year={2018},
  publisher={Nature Publishing Group US New York}
}

@article{reyna2020early,
  title={Early prediction of sepsis from clinical data: the PhysioNet/Computing in Cardiology Challenge 2019},
  author={Reyna, Matthew A and Josef, Christopher S and Jeter, Russell and Shashikumar, Supreeth P and Westover, M Brandon and Nemati, Shamim and Clifford, Gari D and Sharma, Ashish},
  journal={Critical care medicine},
  volume={48},
  number={2},
  pages={210--217},
  year={2020},
  publisher={LWW}
}

@article{johnson2018comparative,
  title={A comparative analysis of sepsis identification methods in an electronic database},
  author={Johnson, Alistair EW and Aboab, Jerome and Raffa, Jesse D and Pollard, Tom J and Deliberato, Rodrigo O and Celi, Leo A and Stone, David J},
  journal={Critical care medicine},
  volume={46},
  number={4},
  pages={494--499},
  year={2018},
  publisher={LWW}
}

@article{stern2023defining,
  title={Defining posttraumatic sepsis for population-level research},
  author={Stern, Katherine and Qiu, Qian and Weykamp, Michael and O’Keefe, Grant and Brakenridge, Scott C},
  journal={JAMA Network Open},
  volume={6},
  number={1},
  pages={e2251445--e2251445},
  year={2023},
  publisher={American Medical Association}
}

@article{eguia2020trends,
  title={Trends, cost, and mortality from sepsis after trauma in the United States: an evaluation of the national inpatient sample of hospitalizations, 2012--2016},
  author={Eguia, Emanuel and Bunn, Corinne and Kulshrestha, Sujay and Markossian, Talar and Durazo-Arvizu, Ramon and Baker, Marshall S and Gonzalez, Richard and Behzadi, Faraz and Churpek, Matthew and Joyce, Cara and others},
  journal={Critical care medicine},
  volume={48},
  number={9},
  pages={1296--1303},
  year={2020},
  publisher={LWW}
}

@article{eriksson:2019-comparison,
    author = {Eriksson, Jesper and Eriksson, Mikael and Brattstr{\"o}m, Olof and Hellgren, Elisabeth and Friman, Ola and Gidl{\"o}f, Andreas and Larsson, Emma and Oldner, Anders},
    journal = {Journal of Critical Care},
    pages = {125--129},
    publisher = {Elsevier},
    title = {Comparison of the sepsis-2 and sepsis-3 definitions in severely injured trauma patients},
    volume = {54},
    year = {2019},
}

@INPROCEEDINGS{9005805,
  author={Morrill, James and Kormilitzin, Andrey and Nevado-Holgado, Alejo and Swaminathan, Sumanth and Howison, Sam and Lyons, Terry},
  booktitle={2019 Computing in Cardiology (CinC)}, 
  title={The Signature-Based Model for Early Detection of Sepsis From Electronic Health Records in the Intensive Care Unit}, 
  year={2019},
  volume={},
  number={},
  pages={Page 1-Page 4},
  doi={10.22489/CinC.2019.014}}

@INPROCEEDINGS{9005773,
  author={Li, Xiang and Kang, Yanni and Jia, Xiaoyu and Wang, Junmei and Xie, Guotong},
  booktitle={2019 Computing in Cardiology (CinC)}, 
  title={TASP: A Time-Phased Model for Sepsis Prediction}, 
  year={2019},
  volume={},
  number={},
  pages={Page 1-Page 4},
  doi={10.22489/CinC.2019.049}}

@INPROCEEDINGS{9005856,
  author={Nonaka, Naoki and Seita, Jun},
  booktitle={2019 Computing in Cardiology (CinC)}, 
  title={Demographic Information Initialized Stacked Gated Recurrent Unit for an Early Prediction of Sepsis}, 
  year={2019},
  volume={},
  number={},
  pages={1-4},
  doi={10.22489/CinC.2019.153}}

@INPROCEEDINGS{9005879,
  author={Liu, Luchen and Wu, Haoxian and Wang, Zichang and Liu, Zequn and Zhang, Ming},
  booktitle={2019 Computing in Cardiology (CinC)}, 
  title={Early Prediction of Sepsis From Clinical Data via Heterogeneous Event Aggregation}, 
  year={2019},
  volume={},
  number={},
  pages={Page 1-Page 4},
  doi={10.22489/CinC.2019.157}}

@INPROCEEDINGS{9005565,
  author={Tran, Luan and Nguyen, Manh and Shahabi, Cyrus},
  booktitle={2019 Computing in Cardiology (CinC)}, 
  title={Representation Learning for Early Sepsis Prediction}, 
  year={2019},
  volume={},
  number={},
  pages={1-4},
  doi={10.22489/CinC.2019.021}}

@article{li2023real,
  title={Real-Time Prediction of Sepsis in Critical Trauma Patients: Machine Learning--Based Modeling Study},
  author={Li, Jiang and Xi, Fengchan and Yu, Wenkui and Sun, Chuanrui and Wang, Xiling and others},
  journal={JMIR Formative Research},
  volume={7},
  number={1},
  pages={e42452},
  year={2023},
  publisher={JMIR Publications Inc., Toronto, Canada}
}

@article{guo2024developing,
  title={Developing an early warning system for detecting sepsis in patients with trauma},
  author={Guo, Kucun and Pan, Bao and Zhang, Xinliang and Hu, Dezheng and Xu, Guangyue and Wang, Lin and Dong, Shimin},
  journal={International Wound Journal},
  volume={21},
  number={1},
  pages={e14652},
  year={2024},
  publisher={Wiley Online Library}
}

@article{rhee2020sepsis,
  title={Sepsis trends: increasing incidence and decreasing mortality, or changing denominator?},
  author={Rhee, Chanu and Klompas, Michael},
  journal={Journal of Thoracic Disease},
  volume={12},
  number={Suppl 1},
  pages={S89},
  year={2020}
}

@inproceedings{fu2019early,
  title={Early sepsis prediction in icu trauma patients with using an improved cascade deep forest model},
  author={Fu, Mengsha and Yuan, Jiabin and Bei, Chen},
  booktitle={2019 IEEE 10th International Conference on Software Engineering and Service Science (ICSESS)},
  pages={634--637},
  year={2019},
  organization={IEEE}
}

@article{horn2022hla,
  title={HLA-A locus is associated with sepsis and septic shock after traumatic injury},
  author={Horn, Dara L and Mindrinos, Michael and Anderson, Kirsten and Krishnakumar, Sujatha and Wang, Chunlin and Li, Ming and Hollenbach, Jill and O’Keefe, Grant E},
  journal={Annals of surgery},
  volume={275},
  number={1},
  pages={203--207},
  year={2022},
  publisher={LWW}
}

@article{rhee2019variation,
  title={Variation in identifying sepsis and organ dysfunction using administrative versus electronic clinical data and impact on hospital outcome comparisons},
  author={Rhee, Chanu and Jentzsch, Maximilian S and Kadri, Sameer S and Seymour, Christopher W and Angus, Derek C and Murphy, David J and Martin, Greg S and Dantes, Raymund B and Epstein, Lauren and Fiore, Anthony E and others},
  journal={Critical care medicine},
  volume={47},
  number={4},
  pages={493--500},
  year={2019},
  publisher={LWW}
}

@article{bosch2022predictive,
  title={Predictive validity of the sequential organ failure assessment score versus claims-based scores among critically ill patients},
  author={Bosch, Nicholas A and Law, Anica C and Rucci, Justin M and Peterson, Daniel and Walkey, Allan J},
  journal={Annals of the American Thoracic Society},
  volume={19},
  number={6},
  pages={1072--1076},
  year={2022},
  publisher={American Thoracic Society}
}

@article{minei2012changing,
  title={The changing pattern and implications of multiple organ failure after blunt injury with hemorrhagic shock},
  author={Minei, Joseph P and Cuschieri, Joseph and Sperry, Jason and Moore, Ernest E and West, Michael A and Harbrecht, Brian G and O’Keefe, Grant E and Cohen, Mitchell J and Moldawer, Lyle L and Tompkins, Ronald G and others},
  journal={Critical care medicine},
  volume={40},
  number={4},
  pages={1129--1135},
  year={2012},
  publisher={LWW}
}

@INPROCEEDINGS{10783411,
  author={Elmerahi, Hadj Ali and Atmani, Baghdad and Barigou, Fatiha and Khemliche, Belarbi and Errouane, Badreddine and Bousmaha, Mohammed},
  booktitle={2024 4th International Conference on Embedded \& Distributed Systems (EDiS)}, 
  title={Parrallel LSTM-DNN Fusion Model for Early Prediction of Sepsis in Intensive Care Units}, 
  year={2024},
  volume={},
  number={},
  pages={43-48},
  doi={10.1109/EDiS63605.2024.10783411}}

@article{lauritsen2020explainable,
  title={Explainable artificial intelligence model to predict acute critical illness from electronic health records},
  author={Lauritsen, Simon Meyer and Kristensen, Mads and Olsen, Mathias Vassard and Larsen, Morten Skaarup and Lauritsen, Katrine Meyer and J{\o}rgensen, Marianne Johansson and Lange, Jeppe and Thiesson, Bo},
  journal={Nature communications},
  volume={11},
  number={1},
  pages={3852},
  year={2020},
  publisher={Nature Publishing Group UK London}
}

@ARTICLE{10701306,
  author={Lee, Seunghee and Shin, Geonchul and Hwang, Jeongseok and Hwang, Yunjeong and Jang, Hyunwoo and Han Park, Ju and Han, Sunmi and Ryu, Kyeongmin and Kim, Jong-Yeup},
  journal={IEEE Access}, 
  title={Early Prediction of Sepsis in the Intensive Care Unit Using the GRU-D-MGP-TCN Model}, 
  year={2024},
  volume={12},
  number={},
  pages={148294-148304},
  doi={10.1109/ACCESS.2024.3470851}}

@article{lauritsen2020early,
  title={Early detection of sepsis utilizing deep learning on electronic health record event sequences},
  author={Lauritsen, Simon Meyer and Kal{\o}r, Mads Ellersgaard and Kongsgaard, Emil Lund and Lauritsen, Katrine Meyer and J{\o}rgensen, Marianne Johansson and Lange, Jeppe and Thiesson, Bo},
  journal={Artificial Intelligence in Medicine},
  volume={104},
  pages={101820},
  year={2020},
  publisher={Elsevier}
}

@inproceedings{teredesai2022sub,
  title={Sub-sequence graph representation learning on high variability data for dynamic risk prediction in critical care},
  author={Teredesai, Ankur and Huang, Sijin and Stewart, Tucker and Hu, Juhua and Thakker, Armaan and Stern, Katherine and O’Keefe, Grant E},
  booktitle={2022 IEEE International Conference on Big Data (Big Data)},
  pages={2082--2092},
  year={2022},
  organization={IEEE}
}

@inproceedings{ewig2023multi,
  title={Multi-subset approach to early sepsis prediction},
  author={Ewig, Kevin and Lin, Xiangwen and Stewart, Tucker and Stern, Katherine and O'Keefe, Grant and Teredesai, Ankur and Hu, Juhua},
  booktitle={2023 Congress in Computer Science, Computer Engineering, \& Applied Computing (CSCE)},
  pages={1335--1341},
  year={2023},
  organization={IEEE}
}

@INPROCEEDINGS{9629559,
  author={Ramos, Guilherme and Gjini, Erida and Coelho, Luis and Silveira, Margarida},
  booktitle={2021 43rd Annual International Conference of the IEEE Engineering in Medicine \& Biology Society (EMBC)}, 
  title={Unsupervised learning approach for predicting sepsis onset in ICU patients}, 
  year={2021},
  volume={},
  number={},
  pages={1916-1919},
  doi={10.1109/EMBC46164.2021.9629559}}

@inproceedings{stewart2023nprl,
  title={NPRL: Nightly Profile Representation Learning for Early Sepsis Onset Prediction in ICU Trauma Patients},
  author={Stewart, Tucker and Stern, Katherine and O’Keefe, Grant and Teredesai, Ankur and Hu, Juhua},
  booktitle={2023 IEEE International Conference on Big Data (BigData)},
  pages={1843--1852},
  year={2023},
  organization={IEEE}
}

@INPROCEEDINGS{10487386,
  author={Ewig, Kevin and Lin, Xiangwen and Stewart, Tucker and Stern, Katherine and O'Keefe, Grant and Teredesai, Ankur and Hu, Juhua},
  booktitle={2023 Congress in Computer Science, Computer Engineering, \& Applied Computing (CSCE)}, 
  title={Multi-Subset Approach to Early Sepsis Prediction}, 
  year={2023},
  volume={},
  number={},
  pages={1335-1341},
  doi={10.1109/CSCE60160.2023.00224}}

@article{chawla2002smote,
  title={SMOTE: synthetic minority over-sampling technique},
  author={Chawla, Nitesh V and Bowyer, Kevin W and Hall, Lawrence O and Kegelmeyer, W Philip},
  journal={Journal of artificial intelligence research},
  volume={16},
  pages={321--357},
  year={2002}
}

@article{johnson2016mimic,
  title={MIMIC-III, a freely accessible critical care database},
  author={Johnson, Alistair EW and Pollard, Tom J and Shen, Lu and Lehman, Li-wei H and Feng, Mengling and Ghassemi, Mohammad and Moody, Benjamin and Szolovits, Peter and Anthony Celi, Leo and Mark, Roger G},
  journal={Scientific data},
  volume={3},
  number={1},
  pages={1--9},
  year={2016},
  publisher={Nature Publishing Group}
}

@misc{mimic_code,
  author       = {MIT Laboratory for Computational Physiology},
  title        = {MIMIC Code Repository},
  year         = {2025},
  howpublished = {\url{https://github.com/MIT-LCP/mimic-code}},
  note         = {Accessed: 2025-09-03}
}

@article{pollard2018eicu,
  title={The eICU Collaborative Research Database, a freely available multi-center database for critical care research},
  author={Pollard, Tom J and Johnson, Alistair EW and Raffa, Jesse D and Celi, Leo Anthony and Mark, Roger G and Badawi, Omar},
  journal={Scientific data},
  volume={5},
  pages={180178},
  year={2018},
  publisher={Nature Publishing Group}
}

@article{guirgis:2016-long,
    author={Guirgis, Faheem W and Brakenridge, Scott and Sutchu, Selina and Khadpe, Jay D and Robinson, Taylor and Westenbarger, Richard and Topp, Stephen T and Kalynych, Colleen J and Reynolds, Jennifer and Dodani, Sunita and others},
    journal={Journal of Trauma and Acute Care Surgery},
    number={3},
    pages={525--532},
    publisher={LWW},
    title={The long-term burden of severe sepsis and septic shock: Sepsis recidivism and organ dysfunction},
    volume={81},
    year={2016},
}

@INPROCEEDINGS{10731539,
  author={Zhang, Sijia and Wang, Zunliang and Zhang, Yuyan and Liu, Songqiao},
  booktitle={2024 8th International Conference on Biomedical Engineering and Applications (ICBEA)}, 
  title={A Reinforcement Learning Approach for Predicting the Onset of Septic Shock Patients with Unfair Bias}, 
  year={2024},
  volume={},
  number={},
  pages={182-187},
  doi={10.1109/ICBEA62825.2024.00041}}

@inproceedings{zerveas2021transformer,
  title={A transformer-based framework for multivariate time series representation learning},
  author={Zerveas, George and Jayaraman, Srideepika and Patel, Dhaval and Bhamidipaty, Anuradha and Eickhoff, Carsten},
  booktitle={Proceedings of the 27th ACM SIGKDD conference on knowledge discovery \& data mining},
  pages={2114--2124},
  year={2021}
}

@article{shyalika2024comprehensive,
  title={A comprehensive survey on rare event prediction},
  author={Shyalika, Chathurangi and Wickramarachchi, Ruwan and Sheth, Amit P},
  journal={ACM Computing Surveys},
  volume={57},
  number={3},
  pages={1--39},
  year={2024},
  publisher={ACM New York, NY}
}
